\documentclass[conference]{IEEEtran}
\IEEEoverridecommandlockouts
\usepackage{cite}
\usepackage{amsmath,amssymb,amsfonts}
\usepackage{algorithmic}
\usepackage{graphicx}
\usepackage{textcomp}
\usepackage{xcolor}

\usepackage{url}
\usepackage{mwe}
\usepackage{array}
\usepackage{booktabs}
\usepackage{authblk}
\usepackage{multirow}
\newcommand{\PreserveBackslash}[1]{\let\temp=\\#1\let\\=\temp}
\newcolumntype{C}[1]{>{\PreserveBackslash\centering}p{#1}}
\newcolumntype{R}[1]{>{\PreserveBackslash\raggedleft}p{#1}}
\newcolumntype{L}[1]{>{\PreserveBackslash\raggedright}p{#1}}
\usepackage[colorinlistoftodos,   textsize=tiny,textwidth=2.2in]{todonotes}

\def\BibTeX{{\rm B\kern-.05em{\sc i\kern-.025em b}\kern-.08em
    T\kern-.1667em\lower.7ex\hbox{E}\kern-.125emX}}
\usepackage{balance}
\begin{document}

\title{On Designing Day Ahead and Same Day Ridership Level Prediction Models for City-Scale Transit Networks Using Noisy APC Data\\
}

\author[1]{Jose Paolo Talusan}
\author[1]{Ayan Mukhopadhyay}
\author[2]{Dan Freudberg}
\author[1]{Abhishek Dubey}

\affil[1]{Vanderbilt University}
\affil[2]{Nashville Metropolitan Transit Authority}


\maketitle

\begin{abstract}
The ability to accurately predict public transit ridership demand benefits passengers and transit agencies. Agencies will be able to reallocate buses to handle under or over-utilized bus routes, improving resource utilization, and passengers will be able to adjust and plan their schedules to avoid overcrowded buses and maintain a certain level of comfort. However, accurately predicting occupancy is a non-trivial task. Various reasons such as heterogeneity, evolving ridership patterns, exogenous events like weather, and other stochastic variables, make the task much more challenging. With the progress of big data, transit authorities now have access to real-time passenger occupancy information for their vehicles. The amount of data generated is staggering. While there is no shortage in data, it must still be cleaned, processed, augmented, and merged before any useful information can be generated. In this paper, we propose the use and fusion of data from multiple sources, cleaned, processed, and merged together, for use in training machine learning models to predict transit ridership. We use data that spans a 2-year period (2020-2022) incorporating transit, weather, traffic, and calendar data. The resulting data, which equates to 17 million observations, is used to train separate models for the trip and stop level prediction. We evaluate our approach on real-world transit data provided by the public transit agency of Nashville, TN. We demonstrate that the trip level model based on Xgboost and the stop level model based on LSTM outperform the baseline statistical model across the entire transit service day.
\end{abstract}

\begin{IEEEkeywords}
component, formatting, style, styling, insert
\end{IEEEkeywords}

\section{Introduction}
Public transportation is a vital component in any modern metropolitan city. Access to reliable forms of public transit have been known to have an impact in many aspects, such improved quality of life, reduced carbon emissions, and have an overall positive effect on social equity. However, even with the availability of public transit, it is not always guaranteed that it is always reliable and accessible. On the contrary, they are more often over-stretched or underdeveloped. As a result, most of the work being done is focused on improving the accessibility and reliability of public transit.

Traditional measures of reliable transit systems include trip frequency, punctuality, and travel time. In response, plenty of work has been done with the goal of improving travel times by identifying and reducing causes of delay~\cite{levinson1983analyzing}. However, an often overlooked element in reliability is the perceived comfort of riders~\cite{echaniz_spatial_2022} which can be seen as the a direct consequence of vehicle occupancy and capacity. A frequently overcrowded bus can prevent potential commuters from even considering public transit. Inversely, consistently low rider demand can be seen as an under utilization of already constrained resources. This duality of public transit is often caused by the agencies' constant struggle with providing increased transit coverage amidst highly heterogeneous ridership demand. 

With the progress of big data, transit authorities now have access to and are able to provide real-time passenger occupancy information for their vehicles. Transit agencies such as the Nashville Metropolitan Transit Authority (MTA) uses Automated Passenger Counter (APC) systems that provides stop-level estimates of passenger boarding and alighting. This information have been integrated by apps such as Transit\footnote{\url{https://transitapp.com/}}, which in addition to allowing potential riders to see an estimated future passenger occupancy, also use crowdsourcing to collect occupancy information from riders onboard in an effort to improve service accuracy. From the perspective of passengers, this helps them choose departure times to match their desired comfort level. For the agencies, this can be a reference for them to optimize their services by allocating resources according to predicted ridership demand. Thus, accurately predicting the maximum occupancy of each vehicle in a public transit system is pivotal in improving perceived reliability, resource optimization, and rider comfort. 

Achieving highly accurate occupancy prediction, however, is a difficult task. There are a number of factors that can affect demand ranging from short high impact factors such as sport events and festivals to long-term factors such as school schedule and season. Additionally, stochastic traffic conditions along the route can cause variation in ridership, further increasing uncertainty. Another issue that can affect prediction is sensor data noise. As with any system that relies on a fleet of sensors and a large database, there are bound to be inconsistencies and errors. This is especially true for APC systems, where passenger boarding and alighting information are recorded using infrared sensors installed on vehicle doors~\cite{patnaik_estimation_2004}. This can lead to erratic and misleading information. This issue brings up the need for data preparation and augmentation to ensure that the data is reliable and useful.

In this paper, we implement an end-to-end framework for predicting occupancy at both the stop and route levels. This ensures that our method can react to both short and long-term changes in the public transit system. We do this by analyzing and combining different spatio-temporal data such as weather, traffic, and APC data to develop a model for bus occupancy. First, we investigate how data can be augmented and merged to provide features that would expose the relationship with bus occupancy. Second, we build different models for bus-stop and transit-route levels. Finally, we demonstrate and compare our approach using actual APC data from the public transit agency of Nashville, TN. The main contribution of this paper is implementing a data cleaning and augmentation method that processes and cleans raw APC data. Raw APC data is often noisy and is faced by different issues. Augmenting and cleaning ensure that data used in training models is valid. We generate passenger occupancy from alighting and boarding information.

\textbf{Organization}: The rest of this paper is organized as follows. In Section~\ref{sec:related_work}, we give an overview of the state-of-the-art in occupancy prediction. In Section~\ref{sec:problem}, we present and formulate the problem. We then discuss in-depth the APC data in Section~\ref{sec:data} and the issues accompanying the dataset. In Section~\ref{sec:results}, we validate our proposed models using real-world data from Nashville, TN. Finally, in Section~\ref{sec:conclusion}, we give our conclusions.

\section{Related Work}
\label{sec:related_work}
In this section we discuss the current state-of-the-art methods used in public transit occupancy prediction.

\subsection{Occupancy Prediction}

Given the importance of public transit and the increasing ubiquity of available vehicle data, research in the field of occupancy prediction, also known as passenger flow or transit demand prediction, has been flourishing. There is a considerable number of work done on understanding and mapping the occupancy level in public transport.

Short-term passenger demand forecasting fall into one of two categories, parametric and non-parametric approaches. Traditionally, parametric approaches such as historical averaging~\cite{smith_comparison_1997} and autoregressive integrated moving average (ARIMA)~\cite{williams_ARIMA_1998} have been used to predict not only demand but traffic flow, travel times and vehicle speed. Ever since it was established, ARIMA has been known to perform well in modeling linear and stationary time series. However, ARIMA's shortcomings in taking into account seasonality and capturing non-linear relationships in data are also well known.

In contrast, non-parametric approaches build a non-linear relationship between the input and output variables without any prior knowledge. These methods gained popularity as consequence of the rapidly increasing availability of data from systems such as Advanced Public Transportation Systems (APTS) and Advanced Traveler Information Systems~\cite{patnaik_estimation_2004}. These techniques have been proven effective at forecasting demand based on data gathered through smart cards~\cite{gong_sd-seq2seq_2020,ouyang_lstm-based_2020}. Toque et al.~\cite{toque_short_2017} used Random Forest (RF) and LSTM neural networks trained on smart card data to predict travel demand. By creating multiple temporal units neural networks (MTUNN) and parallel ensemble neural networks (PENN), Tsai et al.~\cite{tsai_neural_2009} showed that it can outperform predictions based on statistical analysis of historical data.

Incorporating other spatio-temporal dataset such as weather and special events have also been explored. Karnberger et al.~\cite{karnberger_networkwide_2020} considered the effect of exogenous events on public transportation ridership. Meanwhile, Zhou et al.~\cite{zhou_impacts_2017} combined smart data and weather information and found that while riders are more resilient to changes in weather, it still has an effect on the overall demand. Finally, Wood et al. generated models the passenger occupancy and demand at the next-stop/any-stop level based on APC and weather data~\cite{wood_development_2022} and proved that even simpler models such as RF and LSTM provide reliable estimates of future data when trained with historical information if demand patterns are fairly stable.

There has been plenty of work done in the field of public transportation with a special focus on improving reliability through understanding and forecasting passenger demand. However, our work is distinct in three ways. First, our work aims to provide occupancy prediction at both the stop and trip levels separately by forecasting short and long term demand. Second, we work on APC data which is fundamentally different from smart card data, which is the data commonly used by prior work. Smart cards are embedded with integrated circuits enabling it to process information, or in this case, allow for contactless ticketing for riding on mass transit. These cards are much more accurate and complete in their data collection~\cite{MA20131,park_smart_card} due in part they require passengers to swipe after getting on and before getting of the vehicle. In contrast, APC data is much more noisy and introduces far more uncertainty in data collection and processing. Third, we focus on implementing this for the entire public transport system and not on a few select routes.




\section{Problem Statement}
\label{sec:problem}
Based on our conversations with the transit agency, they want to be able to identify particular trips and stops which experience overcrowding. Overcrowding increases the chances of passengers not being able to get on the bus and decreasing their overall satisfaction and willingness to take public transit again in the future. Knowing the maximum occupancy at the trip and stop level will allow them to react and prepare accordingly by increasing bus dispatch frequency thereby decreasing headway.

The primary objective of this work is to provide accurate occupancy prediction for public transit vehicles. The goal is to be able to reliable and efficiently forecast maximum ridership demand at both stop and trip levels. The problem then is, given a fleet of heterogeneous vehicles\footnote{In this work we use the terms vehicle and bus as public transit vehicles interchangeably.}, each equipped with automated passenger count systems, how are we able to model and accurately predict the maximum occupancy at any trip or stop in the future.

\begin{figure}[!htbp]
   \centering
\includegraphics[width=0.9\linewidth]{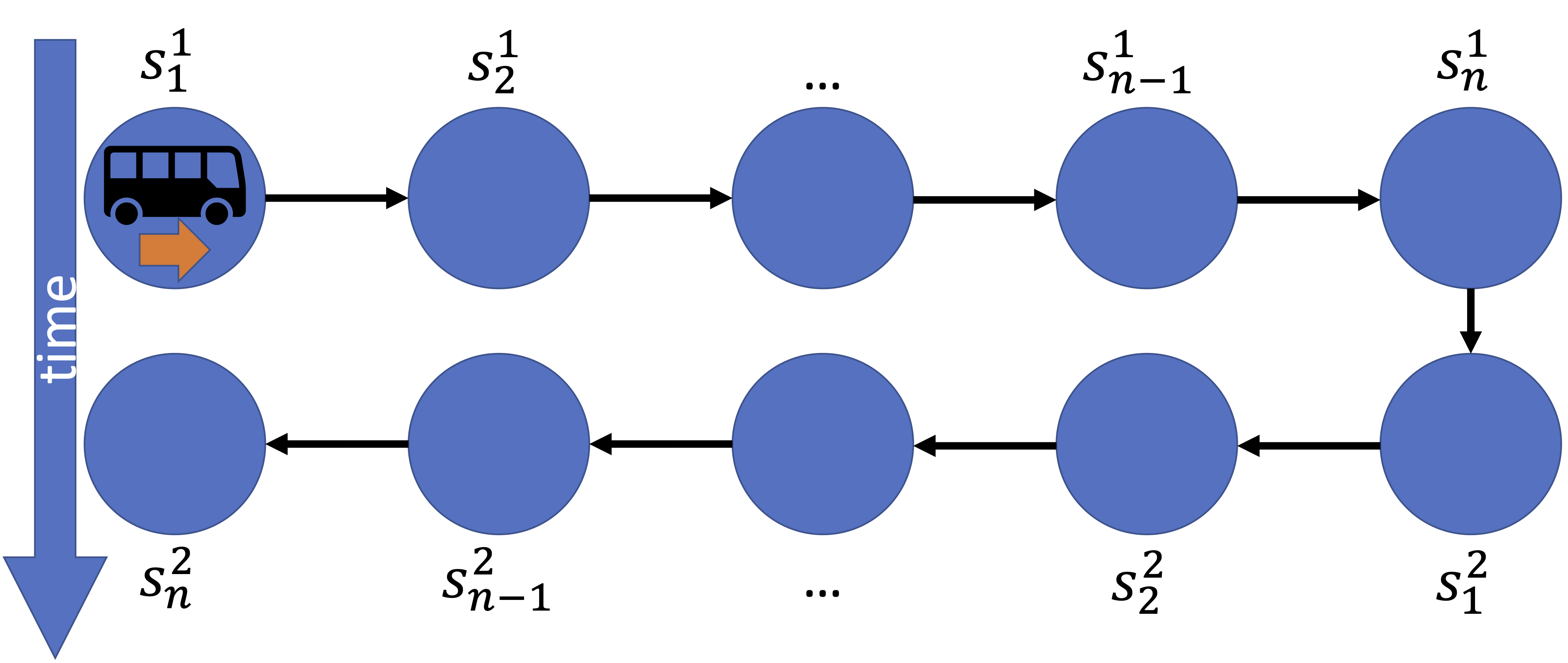}
   \caption{A sample block assignment for a public transit vehicle}
   \label{fig:diagram}
\end{figure}

In a public transit timetable, each vehicle is assigned to serve a specific block. Each block is a collection of non-overlapping trips going back and forth a predetermined route. A single trip $t \in \mathcal{T}$ constitutes a vehicle travelling one direction across a single route, a round trip is made of two separate trips. In each trip, the bus passes by a predetermined number of stops $s \in \mathcal{S}$ where passengers can board from or alight to. At each stop a number of people get on or off the bus, this information is then recorded in the APC data as \texttt{ons} and \texttt{offs}, respectively.

We formally define a bus schedule as a collection of sequential trips $\{t_1, \dots, t_r\}$ assigned to a bus, where each trip $t_r$ is an ordered sequence of $n$ stops $\{s_1^r,\dots,s_n^r\}$. Fig.~\ref{fig:diagram} shows two trips that have been assigned to a bus. The first row of stops $s_1^1$ to $s_{1+n}^1$ correspond to a trip $t_1$ with $n$ stops. Once the vehicle reaches the end of this trip, it proceeds with its return trip, $t_2$, segment of the assigned route. Trip $t_2$ consists of stops from $s_1^2$ to $s_{1+n}^2$. 

Our goal is to predict the passenger occupancy at the stop and trip levels. For the stop level, given a vehicle is at stop $s_1^1$, the goal is to predict occupancy at $s_{2}^1$. For the trip level, the goal is to predict maximum occupancy across the entire trip for any trip in the future, $t_r$. 


\section{Data Collection and Processing}
\label{sec:data}
In this section we first provide an overview of the different data sources used and we describe the data augmentation and processing methods that we applied to it.


\begin{table*}[]
\centering
\caption{Data Features, Size and Sources}
\resizebox{\textwidth}{!}{%
\begin{tabular}{|l|l|l|l|l|l|l|l|l|}
\hline
\textbf{Dataset}          & \textbf{Range} & \textbf{Size}          & \textbf{Rows}                   & \textbf{Features}          & \textbf{Source} & \textbf{Frequency} & \textbf{Type}   & \textbf{Description}                                                                                         \\ \hline
\multirow{12}{*}{Transit} &                 & \multirow{12}{*}{831MB} & \multirow{12}{*}{17,000,000}    & Transit date               & APC             & variable           & Temporal        & Date of bus trip                                                                 \\ \cline{5-9}
                          &                 &                        &                                & Route ID                   & APC             & variable           & Spatio-temporal & Unique route identifier                                                                      \\ \cline{5-9}
                          &                 &                        &                                & Route direction name       & APC             & variable           & Spatio-temporal & Name of route heading                                                         \\ \cline{5-9} 
                          &                 &                        &                                & Scheduled headway          & APC             & variable           & Spatio-temporal & Duration between buses headed in the same route and direction (per stop)           \\ \cline{5-9} 
                          &                 &                        &                                & Load                       & derived         & variable           & Spatio-temporal & Total occupancy at the stop (after alights and boards)                                                       \\ \cline{5-9} 
                          & 01/01/2020      &                        &                                & Stop sequence              & APC             & variable           & Spatio-temporal & Number of current stop within the entire trip                             \\ \cline{5-9} 
                          & to              &                        &                                & Stop ID                    & APC             & variable           & Spatio-temporal & Unique stop identifier                                                        \\ \cline{5-9} 
                          & 04/06/2022      &                        &                                & Past load                  & derived         & variable           & Spatio-temporal & Past loads from previous trips and stops                                    \\\cline{5-9} 
                          &                 &                        &                                & Past actual headway        & derived         & variable           & Spatio-temporal & Past actual headway from previous trips and stops                     \\\cline{5-9} 
                          &                 &                        &                                & Percent load change        & derived         & variable           & Spatio-temporal & Percent change of occupancy from two stops or trips prior                \\\cline{5-9} 
                          &                 &                        &                                & Percent headway change     & derived         & variable           & Spatio-temporal & Percent change of headway from two stops or trips prior                    \\\cline{5-9} 
                          &                 &                        &                                & Zero load at trip end      & APC             & variable           & Spatio-temporal & Boolean indicator if people should all alight at the end of the trip          \\ \hline
\multirow{3}{*}{Weather}  & 01/01/2020      & \multirow{3}{*}{300MB} & \multirow{3}{*}{226,105}       & Temperature                & Darksky         & 1 hour             & Spatio-temporal & Recorded temperature                                                        \\  \cline{5-9} 
                          & to              &                        &                                & Humidity                   & Darksky         & 1 hour             & Spatio-temporal & Recorded humidity                                                            \\  \cline{5-9} 
                          & 04/06/2022      &                        &                                & Precipitation intensity    & Darksky         & 1 hour             & Spatio-temporal & Amount of precipitation.                                                   \\  \hline
\multirow{3}{*}{Traffic}  & 01/01/2020      & \multirow{3}{*}{21GB}  & \multirow{3}{*}{2,300,000,000} & Speed                      & INRIX           & 5 minutes          & Spatio-temporal & Recorded road segment traffic speed                                        \\  \cline{5-9} 
                          & to              &                        &                                &                            &                 &                    &                 &                                                                             \\  \cline{5-9} 
                          & 02/28/2022      &                        &                                &                            &                 &                    &                 &                                                                               \\ \hline
\multirow{3}{*}{Holidays} & 01/01/2020      & \multirow{3}{*}{1MB}   &                                & School breaks              & calendar        & 1 day              & Tmporal         & Scheduled school breaks and holidays                                        \\  \cline{5-9} 
                          & to              &                        &                                & National holidays          & calendar       & 1 day               & Temporal        & National holidays                                                       \\  \cline{5-9} 
                          & 04/06/2022      &                        &                                &                            &                 &                    &                 &                                                                               \\ \hline
\end{tabular}
}
\label{tab:dataset_details}
\end{table*}

\subsection{Data Sources}
There are a variety of data from different sensors and sources that needs to be temporally and spatially joined together.

\begin{itemize}
    \item Automatic Passenger Counting (APC): Automatic Passenger Counting systems record a variety of information as the vehicle passes by bus stops.  Sensors installed over vehicle doors are triggered when people exit and enter the bus, recording \texttt{offs} and \texttt{ons} respectively. Each entry in the APC is a log of the current state of the bus at a stop on a trip. This log also includes scheduled and actual stop arrival times.
    \item Weather: Weather data comes from multiple sources (Darksky and Weatherbit) and multiple weather stations. Data is matched based on the geographic locations of the stops and weather stations, and then joined with the APC data. Data includes precipitation, temperature and humidity.
    \item General Transit Feed Specification (GTFS): A dataset provided by the transit agency. It includes all the schedules and time tables for all the vehicles in their fleet. It also includes the geometric routes and scheduled arrival times that can be used to compute scheduled headways and match with road traffic data.
    \item Traffic: Traffic data is from INRIX~\cite{INRIX}. It provides road segment level speed and congestion information in five minute granularities. Matching this data with APC is done by dividing the metropolitan city into one by one mile grids and identifying the grids where the bus trip's shape passes through. INRIX segments which are within these grids are then collected and the average traffic speed is obtained. This value is then joined with APC.
    \item Calendar: This includes information regarding city holidays and school breaks which have been shown to have an effect in the overall ridership demand~\cite{kashfi_understanding_2015}.
\end{itemize}

\begin{figure}[!htbp]
   \centering
\includegraphics[width=0.9\linewidth]{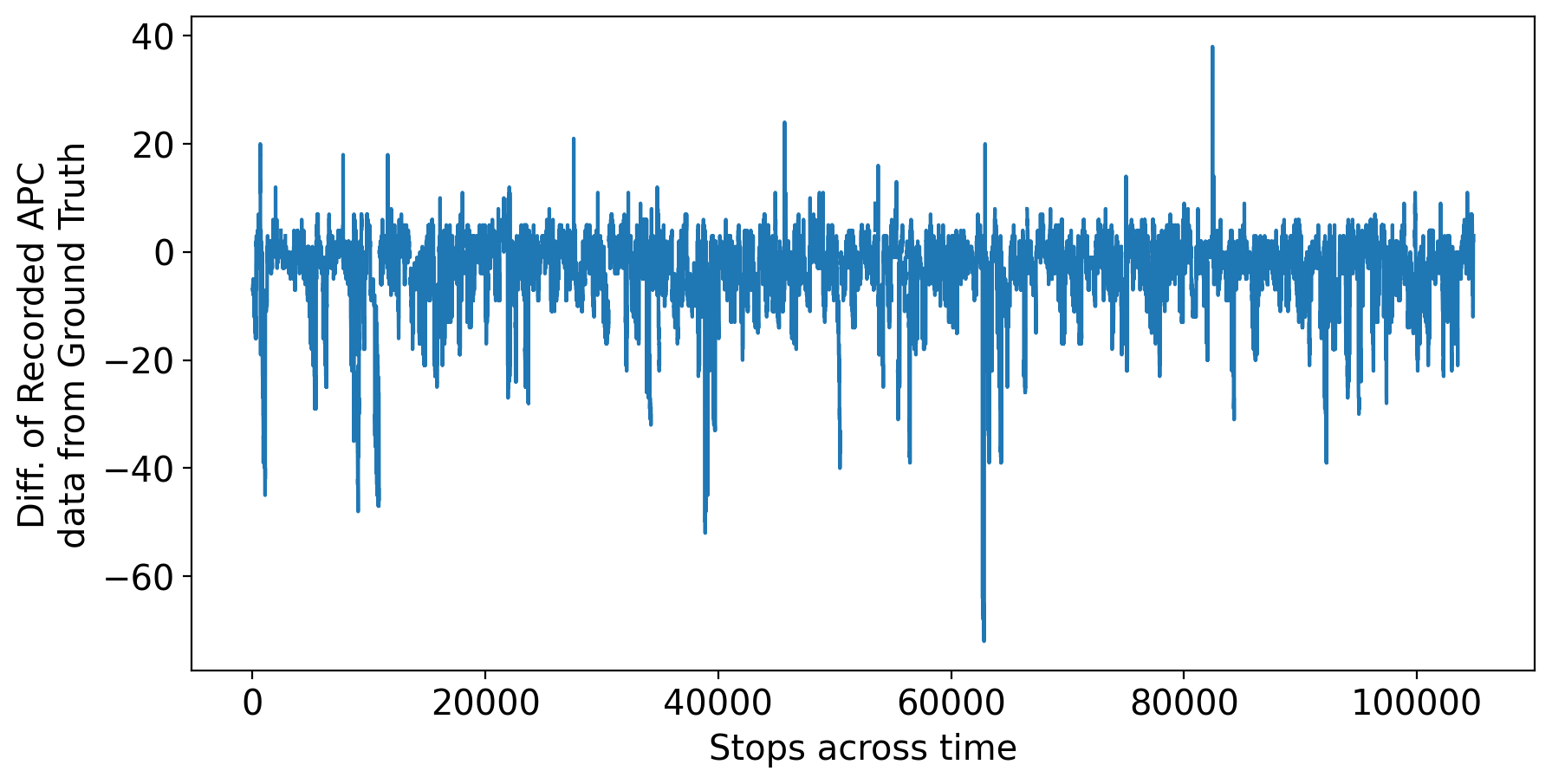}
   \caption{Noise in occupancy sensor readings, difference between ground truth and sensor data observations over a period from 2021-03-21 to 2022-02-17. }
   \label{fig:apc_noise}
\end{figure}

\begin{figure}[!htbp]
   \centering
\includegraphics[width=0.9\linewidth]{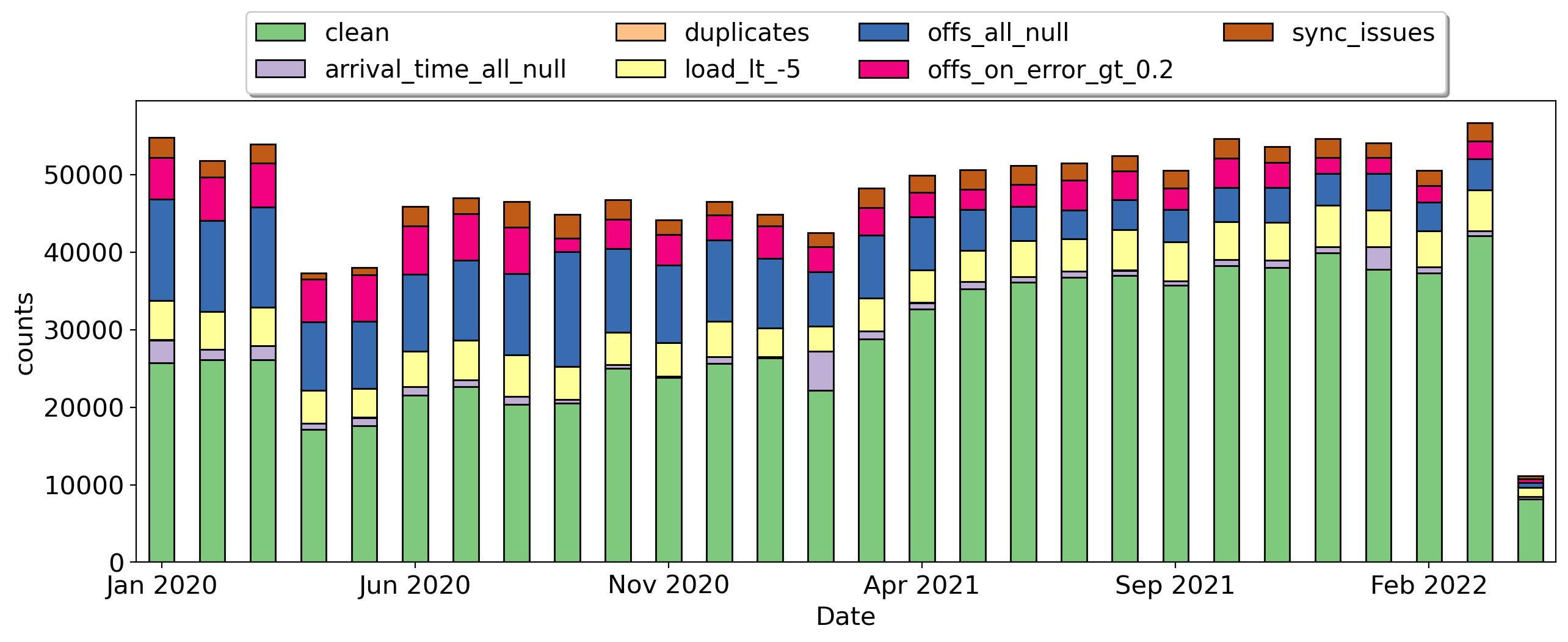}
   \caption{Count of various issues faced when dealing with APC data}
   \label{fig:apc_issues}
\end{figure}

\subsection{Data Cleaning and Augmentation}

APC data received directly from vehicles are noisy, often reporting highly erroneous data. Figure~\ref{fig:apc_noise} shows a plot of the difference between the reported occupancy and ground truth data for a span of almost one year. The recorded data, at average, was 5 people away from the truth, going as much as 72. Another more pressing issue with APC systems is that data is not reliably obtained. Figure~\ref{fig:apc_issues} shows the only a fraction of the data received is in a ``clean" state or a state without issues that could even be attempted to use for training. The issue is continuously improving over time as better APC maintenance practices are implemented, currently there are only less than 20\% of data that is lost and deemed unclean compared to almost 50\% in the past.

Thus, prior to any processing and merging, we filter out trips which have incorrect or unclean entries. Note that while APC data is received at the stop level (events are logged every time a bus reaches a stop), cleaning and filtering are done at the trip level. This is due to the fact that we remove entire trips when any of their stop entries meet certain criteria. The following are the set of rules that determine whether an entry in the APC is valid:

\begin{itemize}
    \item Recorded occupancy is $< -5$.
    \item \texttt{offs} and \texttt{ons} error is $> 0.2$.
    \item All actual arrival times are null.
    \item All \texttt{offs} are null.
    \item The entire trip is a duplicate of a prior trip.
    \item Stop entries are not in the expected chronological sequence, which can sometimes happen when vehicles lose GPS signal.
\end{itemize}

If an entry matches any one of these rules, then it and the entire trip it belongs to are considered invalid and filtered out. Once a valid APC dataset has been established, it is then merged with all the other datasets. Two sets of data are then prepared, one for the trip level and another for the stop level.

Since APC data is recorded every time a bus arrives at a stop, it needs to be aggregated into specific trips before it can be used for trip level occupancy prediction. It is first grouped per \texttt{transit date} and \texttt{trip id}, and aggregated as follows and then be used in trip level occupancy prediction.:
\begin{itemize}
    \item weather: mean weather across all stops since it does not change within the duration of the trip.
    \item headway: mean headway across all stops.
    \item occupancy: maximum occupancy across all stops in the trip.
    \item others: use the first instance as the value.
\end{itemize}

For the stop level prediction, \texttt{zero load at end} an extra feature, which is not present in the trip level data is used. This feature defines whether a trip would require all passengers to alight upon reaching the final stop. The feature is useful for maintaining continuity between trips within a block. Table~\ref{tab:dataset_details} lists down all the features collected and generated from the multiple dataset used in this paper.

Another challenge faced when using APC data for forecast and prediction is the need to sort before any training can begin. In the course of an entire service day multiple vehicles will be travelling across the city, many of the trips occurring simultaneously. Certain blocks are non-overlapping are traversed in sequence by a single vehicle, while others are independent. There might exist multiple routes under each block, each with its own trips that need to be arranged properly before a model such as an LSTM can be used. Otherwise, the data would be disjoint and the model would not be able to learn correctly. 

All of our code is public and available here: \url{https://github.com/smarttransit-ai/mta_occupancy_prediction}

\section{Occupancy Prediction Models}
\label{sec:models}
Recall our goal is to predict passenger occupancy on public transit buses and help transit agencies plan and optimize their trips accordingly. We accomplish this by designing two different models that handle either the stop or trip level ridership demand forecasting. We train and evaluate each model separately. Ultimately, we want to minimize the prediction error for each of the models. Error is measured by how far our model's prediction is from the ground truth.

The ground truth is the occupancy recorded by the APC data. Based on conversations with the transit agency they are interested in primarily identifying trips and stops with a high occupancy count. One of the outputs of this work will be to show potential riders how crowded the arriving bus will be. Thus, a binned output based on the absolute load is sufficient for this problem. We classified the loads based on how the agency breaks it down as well: Low: $\le6$, Medium: $7-12$, Medium-High: $13-54$, High: $55-75$ and Very-High: $\ge76$. However, given the heterogeneity of the buses used in a public transit system, vehicle capacities are not uniform. Thus, using only absolute loads will not provide enough information regarding the crowdedness of a particular bus. One solution should be to factor in the vehicle capacity after inference and provide a crowdedness factor to the user instead of the absolute load. 


\subsection{Feature Selection}
We start with an initial list of 14 features ranging from transit information such as trip date, time and direction to weather and traffic . Features are treated as one of three categories: numerical, one-hot encoded and ordinal. Numerical values include traffic and weather. These values are scaled and normalized before they are used in training. One-hot encoded features include binary features such as is it a holiday, a school break, zero load at end, and also route id and direction and time window. Using one-hot encoding, we can transform these categorical variables into numerical ones while preventing the models from treating one category as greater than the other. On the other hand, we treat year, month, day and hour as ordinal variables where order and sequence are considered. Time windows are not considered ordinal since we want to treat each time window independent of others.

\subsection{Trip Level Prediction}
In trip level prediction, the goal is to be able identify, throughout the service day, which trips in a route experience a high number of occupancy. This allows transit agencies to react and adjust their timetables to future trips that will have a drastic change in demand. Throughout the entire service day, multiple buses will be plying the same trip along the same route and direction. The time between bus dispatch is defined as the headway. We can control the granularity of the data by selecting different time windows with larger time windows grouping together more trips. Grouping by time windows allow the model to provide a prediction for a specific trip at a given time window regardless of which vehicle is present.
We divided trip level prediction into two different models which we call \textbf{day ahead} and \textbf{any day} prediction.

\subsubsection{Day Ahead}
In day ahead, we use data from the prior day (24 hours) to generate additional features and then predict the occupancy level for the trips in the future. If we are trying to predict the occupancy at trip $t_i$ then:
\begin{itemize}
    \item past actual headway percent change of trip $t_{i-2}$ and $t_{i-1}$
    \item past load percent change of trip $t_{i-2}$ and $t_{i-1}$
    \item past average load of trips ${t_{i-P}, \cdots t_{i-1}}$, where $P$ is the number of past trips in the same route and direction.
    \item past average actual headway of trips ${t_{i-P}, \cdots t_{i-1}}$, where $P$ is the number of past trips in the same route and direction.
\end{itemize}

\begin{figure}[!htbp]
   \centering
\includegraphics[width=0.9\linewidth]{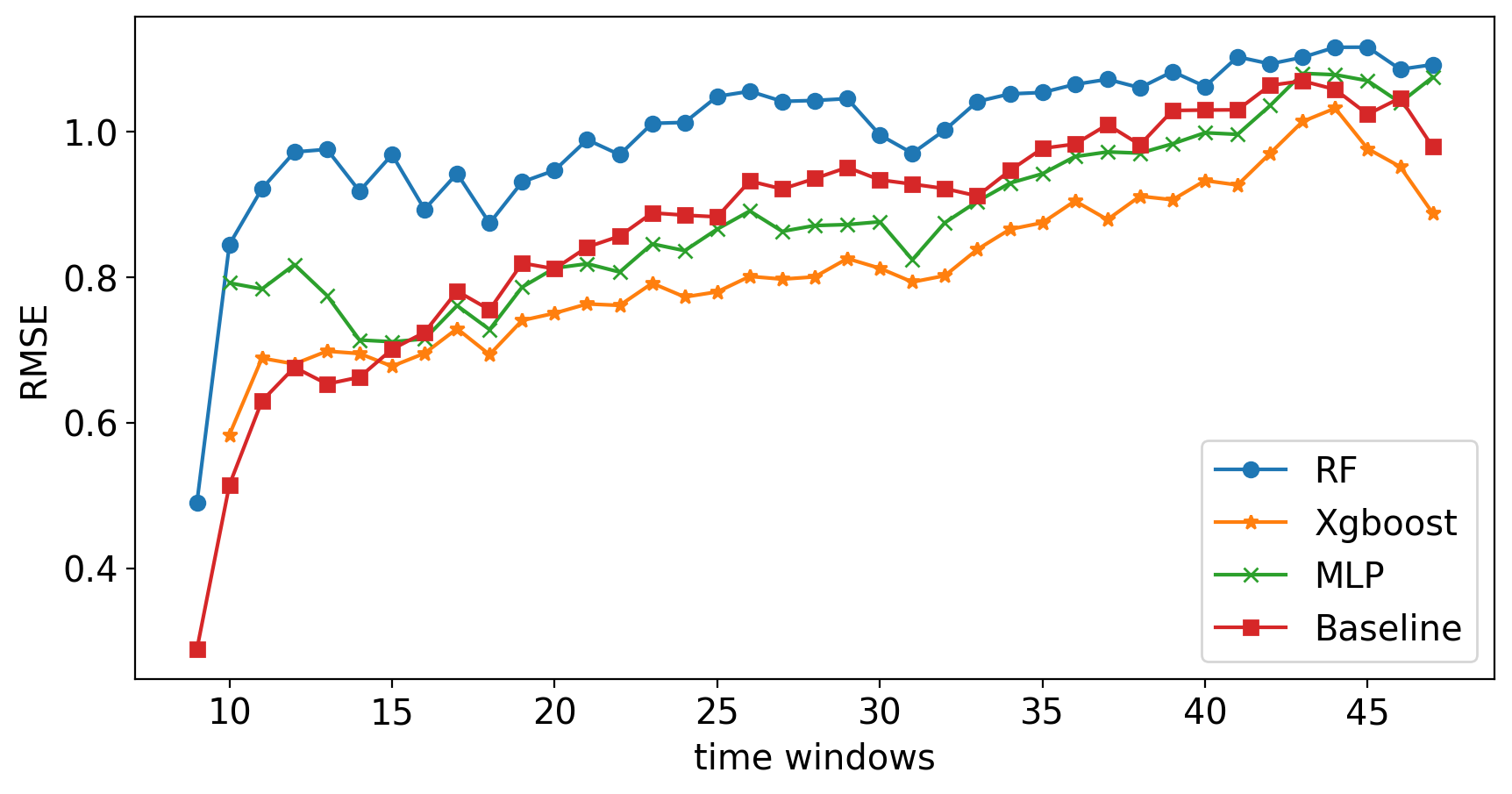}
   \caption{Model comparison for any day prediction. Xgboost RMSE outperforms all others across all time windows.}
   \label{fig:comparison}
\end{figure}

\noindent\textbf{Training} These features along with the features in Table~\ref{tab:dataset_details} are then used as input features in training multiple models ranging from Random Forest, MLP, and an Xgboost model. Figure~\ref{fig:comparison} shows all the RMSE for occupancy prediction across time windows for three models compared to the baseline. Xgboost outperforms all in this preliminary experiment, thus will be used from here on.

\noindent\textbf{Inference} Inference is done on a per trip basis, by providing the transit scheduled for the desired trip $t_{i+1}$, past information, weather and traffic forecast, the output would be the max occupancy of trip $t_{N+1}$. By doing this for all time windows, the transit agency can have an overview of the maximum occupancy at each route and direction across the entire day.

\subsubsection{Any Day Trip Prediction}
This model is used to predict the maximum load occupancy for any trip at any day in the future. This model is similar to the previous model. However, this model does not rely on any past information to generate a prediction. It is trained using the same type of XGBoost model as the day ahead prediction.

\subsection{Stop Level Prediction}
In contrast to the previous two models, stop level attempts to forecast the occupancy at future stops. When used with the trip level prediction, the goal is that it will allow transit agencies to have a more fine-grained view of which stops have a high passenger demand. It uses the stop level dataset generated in Section~\ref{sec:data} as input to our model. The time window is used to group vehicles that travel the same route and direction. 

The data is grouped by transit date, route id, direction, stop id and time window. The occupancy data is then summed across all stops in the same group, giving us an overall idea of the occupancy at that stop for that time window. Similar to the trip level prediction loads are then assigned into the following bins: Low: $\le5$, Medium: $6-11$, Medium-High: $12-16$, High: $17-29$ and Very-High: $\ge30$. The goal of this model is then to predict the binned maximum occupancy for stops ahead given past $p$ stops.

\begin{figure}[!htbp]
   \centering
\includegraphics[width=0.75\linewidth]{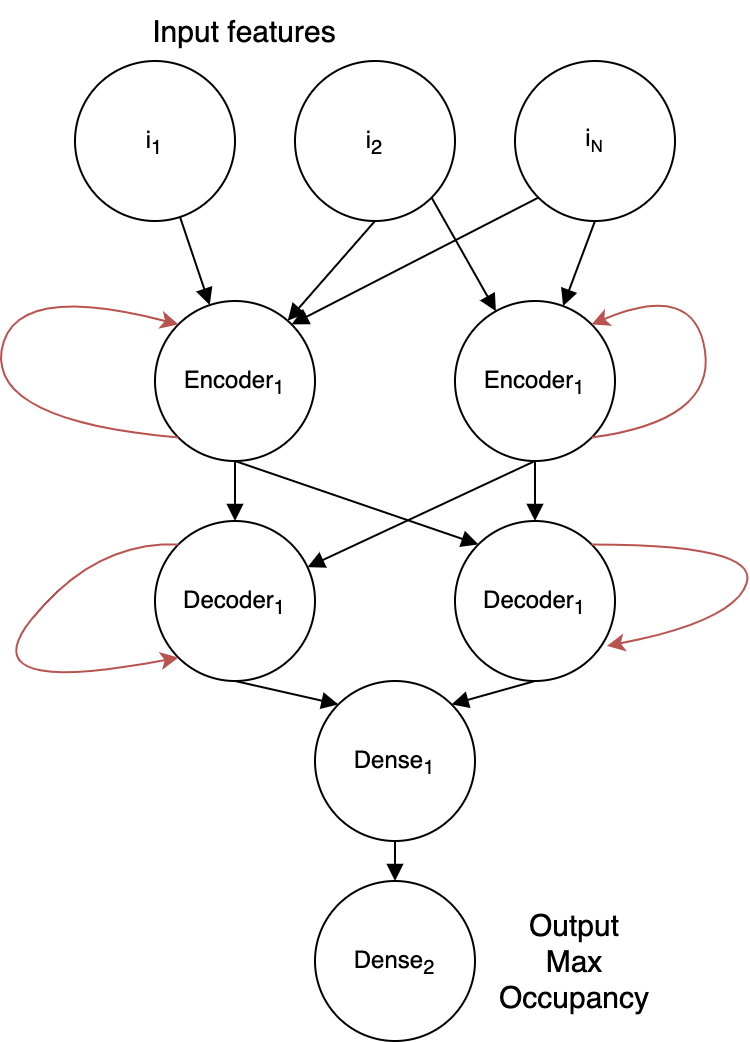}
   \caption{LSTM Encoder-Decoder architecture}
   \label{fig:lstm_architecture}
\end{figure}

\noindent{\textbf{Training:}} Our proposed machine learning model is an LSTM encoder-decoder, see Figure~\ref{fig:lstm_architecture}, with two recurrent modules and two feed forward modules which output a bin corresponding to the maximum occupancy at the stop. An encoder-decoder or seq2seq model is selected to be able to leverage its ability of transforming input into some latent space and using a decoder to create a sequence from those inputs. In essence we are using the model to generate the next word in a sentence, where words are stops and the sentence are the trips. The model is trained on the past stops which are then used to predict the immediate next stop.

\noindent\textbf{Inference:} We use the past N stops to predict the next stop ahead in a single trip. Weather and traffic forecast are combined with scheduled transit date time for use in the prediction. The output is the predicted occupancy at the stop in a particular route and direction.

\section{Results and Discussions}
\label{sec:results}
\begin{figure*}[!htbp]
  \includegraphics[width=\textwidth]{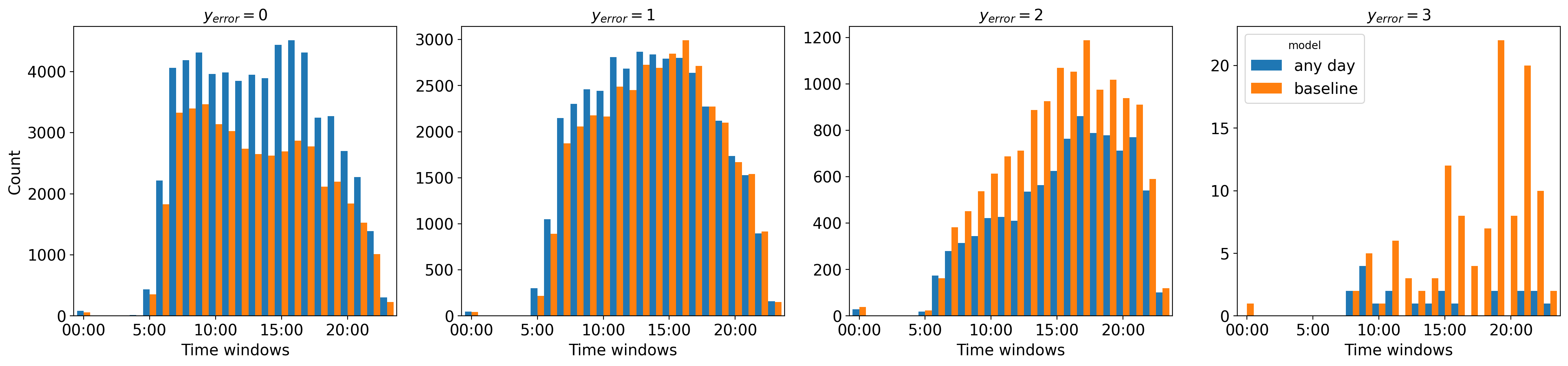}
  \caption{Comparison between baseline and any day trip level models. From left to right, $y_{error} = \{0, 1, 2, 3\}$. The model is able to provide 40\% more correct predictions and 29\% less mistakes than the baseline.}
  \label{fig:triplevel_compare_30minute}
\end{figure*}

In this section, we evaluate our models based on real-world public transit data from Nashville, TN. We describe our experimental setup and then present the results for the trip level and then stop level predictions.

\subsection{Experimental Setup}
We use APC data for Nashville, TN provided by Nashville Metropolitan Transit Authority (MTA). We used 28 months of data from January 2020 to April 2022. Across these two years, MTA has an average of 100 unique vehicles, serving 30 routes going in 10 different directions in a single service day (counting both weekdays and weekends and holidays). In this work, we used all possible route and direction combinations present in the dataset.
All training and experiments were done on a machine with 16-core AMD CPU and 4 Nvidia Titan Xp.
We measure error as the distance between the ground truth and predicted occupancy. We treat the binned classes as ordinal thus, we use:
\begin{equation}
    y_{error} = y_{true}-\hat{y}
\end{equation}
Predictions that are far from the truth have larger errors than predictions that are off by a single bin.

\subsection{Trip Level Prediction}

For the trip level prediction, we split the 430404 trip data into 70\% training and 30\% testing. Trip level prediction model is generated using multiple algorithms such as Random Forest, MLP, LSTM, and XGBoost. Each model is scored based on a 5-fold cross-validation and compared with a baseline model. The baseline model used statistical analysis on historical data. We looked at the past trips taken along the same route and direction, then we get the maximum occupancy across all of those past trips which is then binned. We did this for all trips one, two and four weeks in the past. We found that accuracy does not improve across different baselines. The XGBoost model performs the best compared to all the other models and the baseline.

\begin{figure}[!htbp]
   \centering
\includegraphics[width=0.9\linewidth]{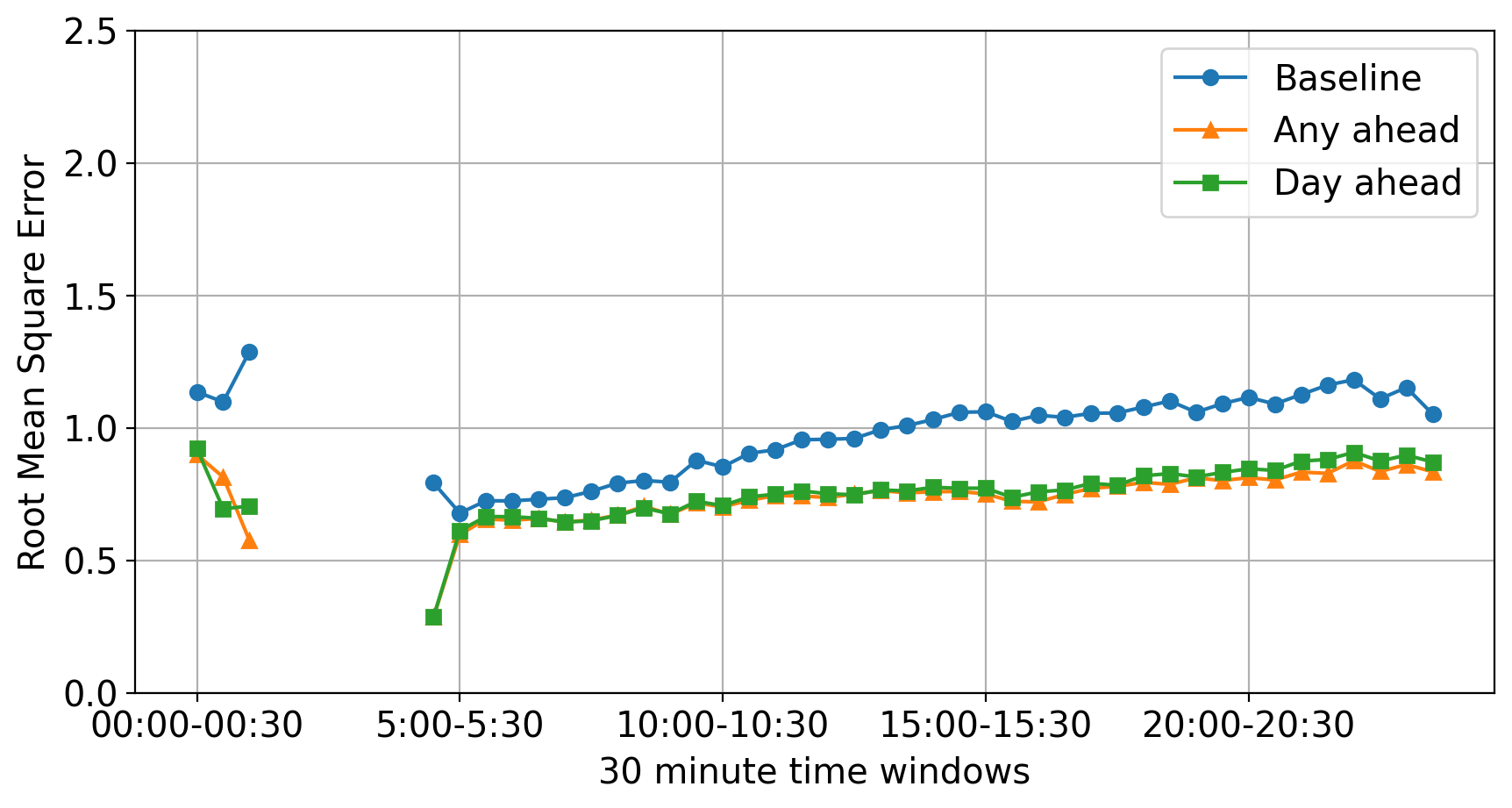}
   \caption{RMSE of baseline, day ahead, and any day models. The gap in the data is due to buses not travelling at those hours (1:30 am to 4:00am).}
   \label{fig:trip_level_overview}
\end{figure}

Grid search based on a 5-fold cross-validation is done to select the best hyperparameters for the model. We tested different time windows for aggregation and at every time window, the model performed better than the baseline. Figure~\ref{fig:triplevel_compare_30minute} shows the counts of errors by time window for the baseline and day ahead model. A higher count of 0 errors and lower count of other errors is preferred. Our model is able to predict more trips correctly across all time windows. The model provides 40\% more correct predictions and makes 29\% less $y_{error} = 2$ mistakes.
Figure~\ref{fig:trip_level_overview} shows the RMSE of each model across time windows. This further proves that our models perform better, however, it also proves that even with the past information included in the day ahead prediction models, it only performs marginally better than the any day model. Note that the gap is due to the buses being unavailable at those hours (1:30 am to 4:00 am).

To understand which features have an effect on the final prediction models, we generated a SHAP analysis for the day ahead model. It shows that the feature with the highest impact on the model output is route and direction. This is expected since certain routes experience more demand than others due simply to the fact that these routes feature destinations that expect a lot of commuters. The next highest ones are hour and month which are due to jobs and schools having a direct effect in demand. Aside from past trip loads, all other past information had little to no impact to the overall model output. Certain features such as school breaks and national holidays also had less impact since these features essentially have the same relationship with transit demand as month.

\begin{figure}[!htbp]
   \centering
\includegraphics[width=0.9\linewidth]{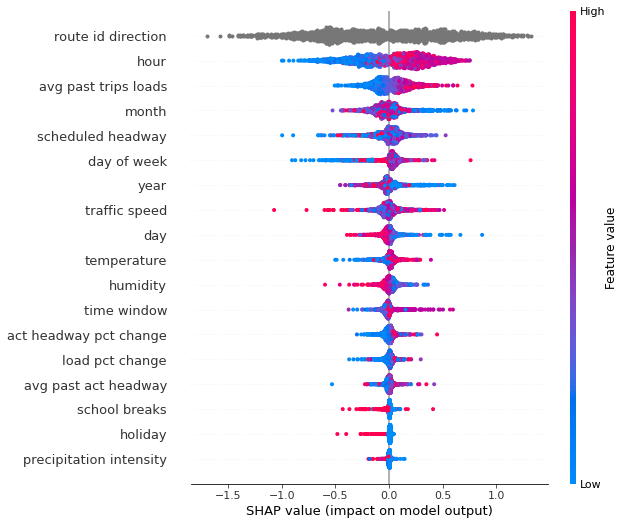}
   \caption{SHAP Feature Analysis for the trip level model}
   \label{fig:SHAP_beeswarm}
\end{figure}

\begin{table}[]
\scriptsize
\centering
\caption{Effect of varying time window on any day prediction of low vs. high occupancy}
\begin{tabular}{@{}cccc@{}}
\toprule
time window (min) & precision & recall & F1 score \\ \midrule
1                 & \textbf{0.5860}    & 0.6289          & 0.6066            \\
10                & 0.5833             & 0.6357          & 0.6083            \\
20                & 0.5750             & 0.6511          & 0.6107            \\
30                & 0.5693             & 0.6615          & 0.6120            \\
40                & 0.5526             & 0.6884          & 0.6131            \\
50                & 0.5469             & 0.6989          & \textbf{0.6136}   \\
60                & 0.5419             & 0.7064          & 0.6133            \\
120               & 0.5043             & \textbf{0.7695} & 0.6093            \\ \bottomrule
\end{tabular}
\label{tab:f1_score_any_day}
\end{table}

Since one of the end goals of this work is to help the transit agency plan for sudden high occupancy events, we evaluate the ability of the model to distinguish between low (0-11) and high (12-100) number of occupants. In Table~\ref{tab:f1_score_any_day} we summarize the precision, recall and F1 scores for the any day prediction given different time windows. The model is able to distinguish high and low occupancy 61\% of the time. We can see the effect of increasing the time window has on precision and recall which is expected since having smaller time windows result in finer grained predictions which can approximate the ground truth better. While the differences in F1 scores are negligible, the most accurate time windows are those between the extremes.

\begin{table}[]
\scriptsize
\centering
\caption{Effect of varying past stops with constant time window (15 minutes) on low vs. high occupancy prediction of next stop}
\begin{tabular}{@{}cccc@{}}
\toprule
past stops & precision & recall & F1 score \\ \midrule
1      & 0.9276    & 0.9435  & 0.9355            \\
3      & \textbf{0.9697}    & 0.9381 & \textbf{0.9536}            \\
5      & 0.9407    & \textbf{0.9623}  & 0.9514           \\
10     & 0.9428    & 0.9569  & 0.9498          \\ \bottomrule


\end{tabular}
\label{tab:f1_score_same_day_past}
\end{table}

\begin{table}[]
\scriptsize
\centering
\caption{Effect of varying time windows stops with constant past stops (5 stops) on low vs. high occupancy prediction of next stop}
\begin{tabular}{@{}cccc@{}}
\toprule
time window & precision & recall & F1 score \\ \midrule
15      & 0.9008 &	0.9077 &	0.9042	\\
30      & 0.9273 &	0.8870 &	0.9067  \\
45      & 0.9783 &	0.9000 &	0.9375  \\
60      & \textbf{0.9881} &	0.9540 &	\textbf{0.9708} \\
90      & 0.9583 &	\textbf{0.9583} &	0.9583 \\ \bottomrule

\end{tabular}
\label{tab:f1_score_same_day_time_window}
\end{table}

\subsection{Stop Level Prediction}

For stop level prediction, we split 17M rows of data into the following:
\begin{itemize}
    \item Training: 2020-01-01 to 2021-06-30
    \item Validation: 2021-06-30 to 2021-10-31
    \item Testing: 2021-10-31 to 2022-04-06
\end{itemize}
A hyperparameter search was done to identify the optimal learning rate, batch size, size of hidden layers, and number of past stops to use in predicting the next stop. The model is compared to multiple baseline models, a simple rolling baseline where only the immediate past stop occupancy is used, a statistical analysis based baseline which gets the max or mean occupancies of the stop in the past (matching route, direction, time window and day of week). Both the number of errors and root mean squared error were used to evaluate the model.

Similar to the any day model, we choose various values of past stops and time windows. We see in Tables~\ref{tab:f1_score_same_day_past} and~\ref{tab:f1_score_same_day_time_window} that using different hyperparameters had an effect on the prediction ability of the model. However, the difference between the values are very small and almost negligible.

\begin{figure}[!htbp]
    \centering
    \begin{minipage}{0.48\linewidth}
        \includegraphics[width=\linewidth]{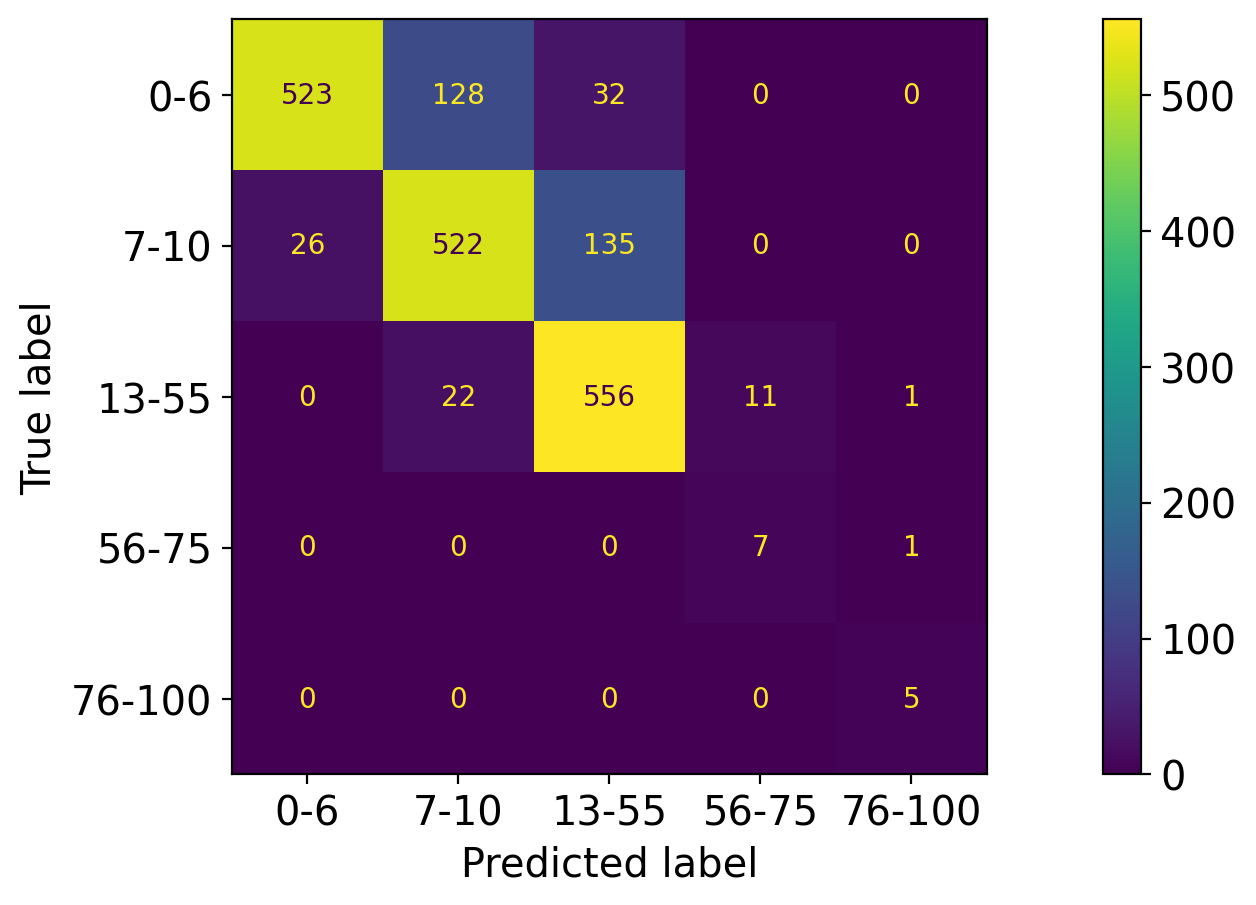}
    \end{minipage}\hfill
    \begin{minipage}{0.48\linewidth}
        \includegraphics[width=\linewidth]{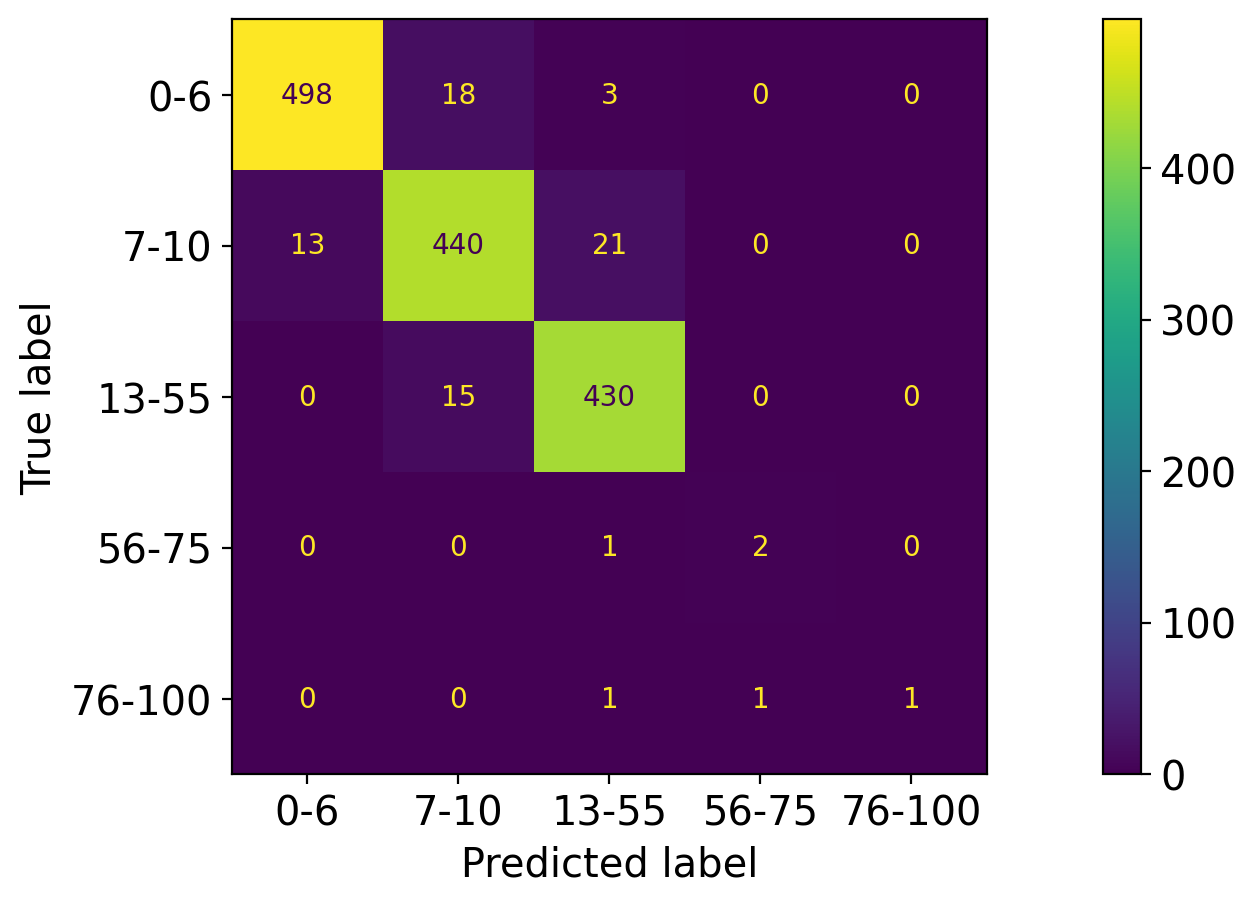}
    \end{minipage}\hfill
    \caption{Confusion matrices when predicting the 6\textsuperscript{th} stop (left) using a rolling baseline, (right) using the past 5\textsuperscript{th} stops as input to the model.}
    \label{fig:confusion_matrix}
\end{figure}

For evaluation we uniformly select 5000 random trips which have at least 10 stops in a trip. We then compare the baselines with the model trained with the hyperparameters resulting from the grid search. Figure~\ref{fig:confusion_matrix} shows the ability of the model to predict the 6\textsuperscript{th} stop given the preceding 5. It is able to predict more accurately than even the rolling baseline.

\begin{figure}[!htbp]
   \centering
\includegraphics[width=0.9\linewidth]{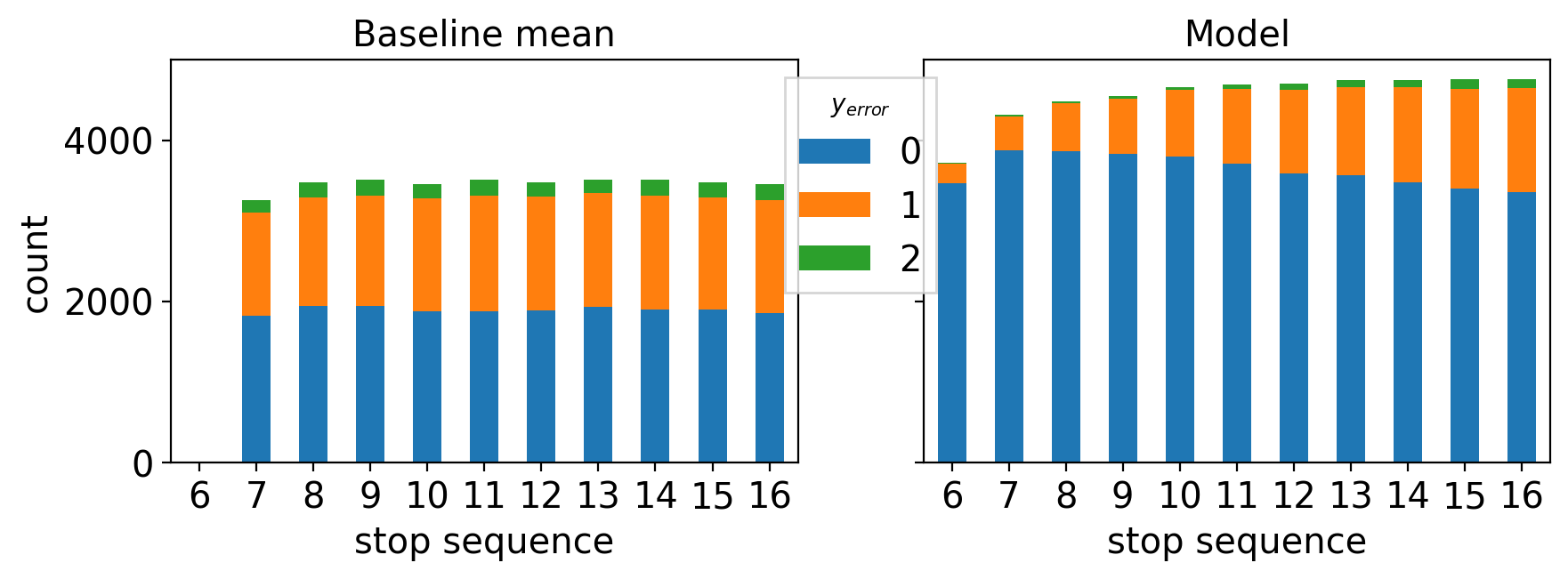}
   \caption{Count of errors per stop in the future}
   \label{fig:sameday_error_count}
\end{figure}

In contrast to the rolling baseline which can only predict the next stop, our model is able to predict any number stops given an initial seed of past stops. Figure~\ref{fig:sameday_error_count} shows the error count as the number of predicted stops increase. The baseline mean is unable to generate a prediction for stops in the future. The difference in counts is due to not being able to find past data that matches the features of the future stop.

Finally we show that the results from the model stay consistent throughout different months.

\begin{figure}[!htbp]
   \centering
\includegraphics[width=0.9\linewidth]{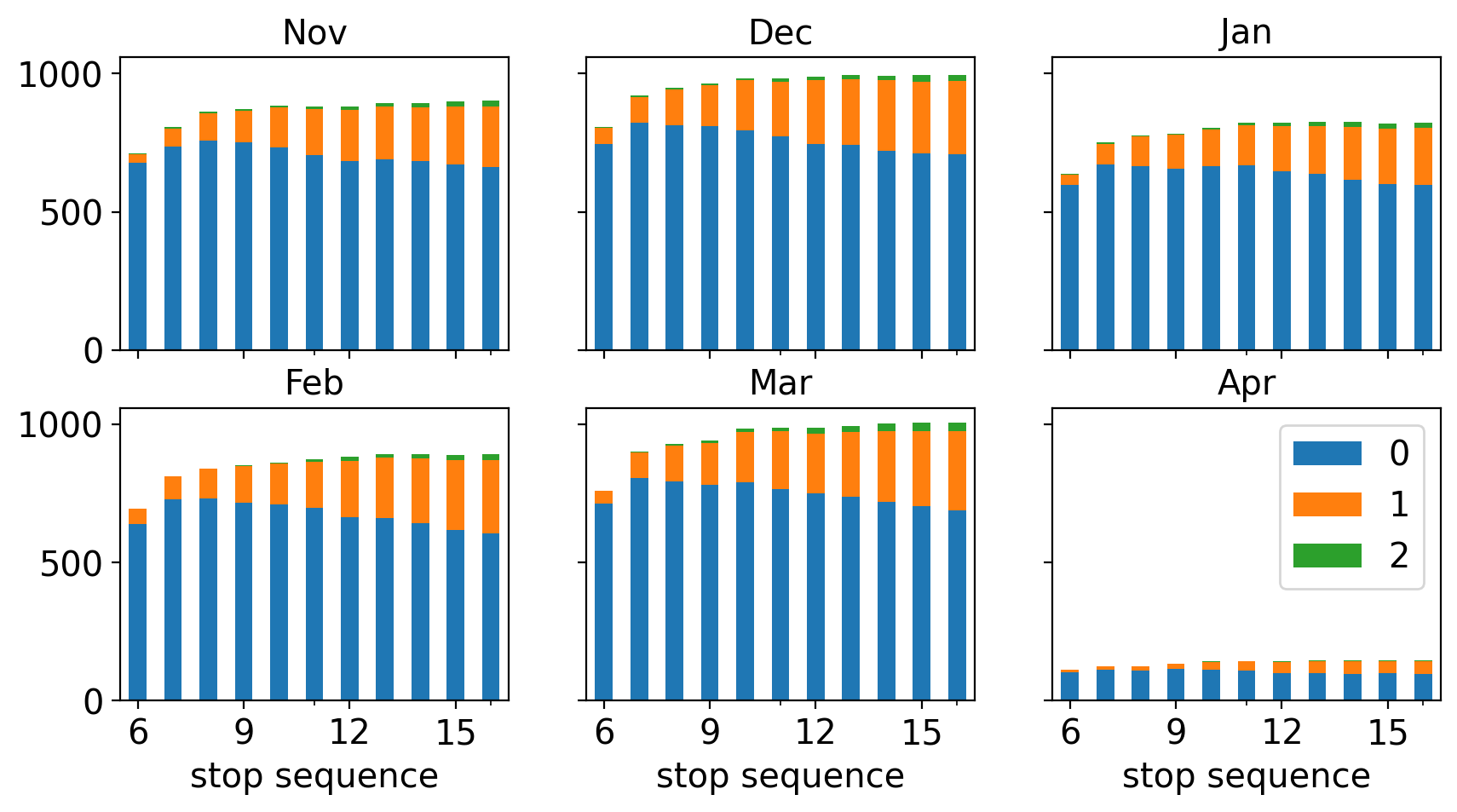}
   \caption{Count of errors per month (2022) per stop in the future}
   \label{fig:sameday_error_count_monthly}
\end{figure}

\section{Conclusion}
\label{sec:conclusion}
The ability to predict and forecast transit occupancy accurately is a boon not only to passengers but to transit agency. Passengers will be able to adjust their schedules or plans to meet their comfort requirements. Transit planners will have the opportunity to allocate resources much more efficiently. However, predicting occupancy is a non-trivial task. Due to the difficulty of this proble, we propose to utilize not only available data from the transit agencies (APC) but to leverage any additional datasets that can provide further insight in the ridership demands. In this paper, we presented a way to collect, process and augment data from the transit agency, and merge it with traffic, weather, GTFS to obtain some meaningful compilation of data. Our key contribution is proposing two separate models for predicting the trip and stop level occupancy. We found that we are able to outperform baseline statistical analysis using the trained models.
\balance
\bibliographystyle{IEEEtran}
\bibliography{my_zotero,extra}
\end{document}